\documentclass[10pt, a4paper]{article}

\usepackage[final]{lrec2026} 
\usepackage{multicol, multirow, booktabs, adjustbox, makecell, tabularx, graphicx, subcaption, xspace}
\usepackage[section]{placeins} 

\newcommand{\intvls}{InternVL3-1B\xspace}
\newcommand{\intvlb}{InternVL3-8B\xspace}
\newcommand{\qwens}{Qwen2.5-3B\xspace} 
\newcommand{\qwenb}{Qwen2.5-7B\xspace}
\newcommand{\llava}{llava-1.5-7b\xspace}
\newcommand{\gpt}{gpt-4o\xspace}
\newcommand{\gptmini}{gpt-4o-mini\xspace}

\title{Predicting Sentence Acceptability Judgments \\ in Multimodal Contexts}

\name{Hyewon Jang\textsuperscript{1}, Nikolai Ilinykh\textsuperscript{1}, Sharid Loáiciga\textsuperscript{1},  Jey Han Lau\textsuperscript{2}, Shalom Lappin\textsuperscript{1,3,4}} 

\address{\textsuperscript{1}University of Gothenburg, \textsuperscript{2}University of Melbourne, \\\textsuperscript{3}Queen Mary University of London, \textsuperscript{4}King's College London\\
         \{hyewon.jang, shalom.lappin\}@gu.se\\}

\abstract{
Previous work has examined the capacity of deep neural networks (DNNs), particularly transformers, to predict human sentence acceptability judgments, both independently of context, and in document contexts. We consider the effect of prior exposure to visual images (i.e., visual context) on these judgments for humans and large language models (LLMs). Our results suggest that, in contrast to textual context, visual images appear to have little if any impact on human acceptability ratings. However, LLMs display the compression effect seen in previous work on human judgments in document contexts. Different sorts of LLMs are able to predict human acceptability judgments to a high degree of accuracy, but in general, their performance is slightly better when visual contexts are removed. Moreover, the distribution of LLM judgments varies among models, with Qwen resembling human patterns, and others diverging from them. 
LLM-generated predictions on sentence acceptability are highly correlated with their normalised log probabilities in general. However, the correlations decrease when visual contexts are present, suggesting that a higher gap exists between the internal representations of LLMs and their generated predictions in the presence of visual contexts. Our experimental work suggests interesting points of similarity and of difference between human and LLM processing of sentences in multimodal contexts.
 \\ \newline  }

\begin{document}

\maketitleabstract

\section{Introduction}
\label{sec:introduction}
There has been a considerable amount of work on using DNNs to predict human sentence acceptability judgments \cite{Lau&etal2017, Bernardy&etal2018,warstadt-etal-2019-neural,Lau&etal2020,qiu&etal2024}. Some previous studies \citep{Bernardy&etal2018,Lau&etal2020} consider the effect of document (textual) context on human acceptability judgments. We examined the impact of visual contexts on both human and LLM sentence ratings.

To prevent data contamination, we selected recent text from the Internet in a variety of genres. We follow \citet{Lau&etal2020} in performing round trip machine translation on sentences, through several languages, using an older statistical MT system, \emph{Moses}, to introduce a variety of syntactic, semantic, and lexical infelicities into the English output. 

We employed Prolific\footnote{\url{https://www.prolific.com/}} crowd sourcing to obtain native speaker human ratings on these sentences, together with a subset of the original English sources, in the context of preceding visual images. GPT-5 generated images for the original English versions of the sentences in our test data. We looked at contexts where the images were relevant to the content of the sentences, contexts where they were irrelevant, and null contexts where no visual images appear. We then tested  closed LLMs and open-source LLMs to predict the human ratings in these contexts. 

Previous work reports a compression effect for human ratings in document contexts, relative to null contexts \citep{Bernardy&etal2018,Lau&etal2020}. This involves raising acceptability at the lower end of the acceptability scale, and lowering them at the higher end. Two explanations have been suggested for this effect, one being due to cognitive load, and the other being due to discourse coherence effect. In the cognitive load account, acceptability ratings gather around the middle because humans experience cognitive load when faced with more input to process, making them more conservative in their judgment. In the discourse coherence account, the ratings for bad sentences are raised because they look less bad when placed in the relevant discourse. In our experiments, we observe no such effect for human acceptability judgments in visual contexts, although there is some indication of a slight raising effect at the lower end, in relevant visual image contexts. We discuss the implications of this finding in \S\ref{sec:humans}.

Four of the seven LLMs that we tested score well on predicting human sentence ratings (0.8-0.9 for Spearman correlations). However, in contrast to humans, two out of four of these models exhibit a clear compression effect for the visual contexts, which closely resembles the one reported in \citet{Lau&etal2020} for human judgments in document contexts (see the three regression graphs in Figure~\ref{fig:lau-etal20}). Overall, our models' predictions are much higher in the absence of visual contexts. We also find varying patterns in the distribution of LLM judgments, with only one of them, \qwenb, resembling the human rating clusters to a degree.

Our experiments suggest that while current LLMs have greatly improved in their ability to identify levels of sentence acceptability relative to earlier DNNs, including first generation transformers, they process sentences in multimodal contexts differently than humans do.

\section{Related Work}

Prior work employs different settings for sentence acceptability judgment. \citet{Lau&etal2017,warstadt-etal-2019-neural,qiu&etal2024} use DNNs to predict human sentence acceptability ratings independently of context. \citet{Lau&etal2020} address the effect of document contexts on sentence acceptability judgment by both humans and DNNs. They employ various probabilistic methods to approximate acceptability judged by DNNs, such as applying normalising score functions to raw probabilities to filter out the effects of lexical frequency and sentence length. 

\citet{Lau&etal2020} crawl natural sentences from Wikipedia and introduce infelicities by round-trip machine translation. A different approach is adopted by \citet{warstadt-etal-2019-neural}, who construct a test corpus (CoLA) consisting largely of linguists' example sentences in minimal pairs, with ratings given by linguists. 

We follow  \citet{Bernardy&etal2018, Lau&etal2020} in using naturally occurring text (modulated through round trip MT), crowd sourced non-expert human rating, and assessment relative to context. \citet{Bernardy&etal2018, Lau&etal2020} examine the effect of preceding document contexts on human sentence acceptability judgments.
We focus on the impact of visual image contexts on both human and LLM acceptability ratings. We also consider the similarities and differences in rating distributions between humans and models. 

More recent studies have been reported that are relevant to our approach in this work. \citet{qiu&etal2024} apply GPT-3 to \citet{Lau&etal2017}'s human rated sentences. They show that it performs well in predicting these ratings. But they do not address the possibility of contamination in which GPT-3 may have been trained on some of this data, which has been publicly available since 2017. To minimise the influence from data contamination we include sentences published online after the training dates of the models for acceptability prediction. Furthermore, \citet{qiu&etal2024} only report results based on prompting experiments. Our work provides more robust findings following \citet{kauf-etal-2024-log}, who compare prompting and logprobs for human judgments of semantic plausibility. They report that logprobs are a good predictor of these judgments.

\section{Human Acceptability Judgments}
\label{sec:humans}

\begin{table*}
    \centering
    \begin{tabular}{ccp{12cm}}
    \toprule
       \multirow{4}{*}{Book} & \multirow{2}{*}{Orig}  &  One piece of felt, folded in the corner, will be spliced, unfurled and dangled from the ceiling like a canopy. \\
       \cmidrule{2-3}
    & \multirow{2}{*}{Mod} & A piece in a corner, spliced were put forward and believe that the upper ceiling as a canopy. \\
    \midrule
    \multirow{4}{*}{News} & \multirow{2}{*}{Orig} &  But the answer seems to be no: Reeves lets it be known she requests no costings on raising the three forbidden taxes.  \\ 
    \cmidrule{2-3}
    & \multirow{2}{*}{Mod} & But the reply does not seem: reeves suggests that it would not cost to raise the three banned taxes.  \\ 
    \midrule
    \multirow{4}{*}{Wiki} & \multirow{2}{*}{Orig} &  In response, the US deployed an additional 170,000 troops during the 2007 troop surge, which helped stabilize parts of the country. \\ 
    \cmidrule{2-3}
    & \multirow{2}{*}{Mod} & In response, we deployed with further 170 000 soldiers in 2007 increase troops, which help to stabilize parts of the country. \\
    \bottomrule
    \end{tabular}
    \caption{Examples of sentences used in our experiment, sourced from books, news, and Wikipedia in original form (Orig) and modified (Mod) through round-trip machine translation with Moses 3.0.}
    \label{tab:sentence-examples}
\end{table*}

\begin{table*}[h]
    \centering
    \begin{tabular}{ccc}
    \toprule
    \textbf{Original Sentence} & \textbf{Relevant Image} & \textbf{Irrelevant Image} \\
    \midrule
    \makecell[c]{One piece of felt, folded in the corner, \\ will be spliced, unfurled and dangled \\ from the ceiling like a canopy.}  & \adjustbox{valign=c}{\includegraphics[width=3.5cm, height=3.5cm]{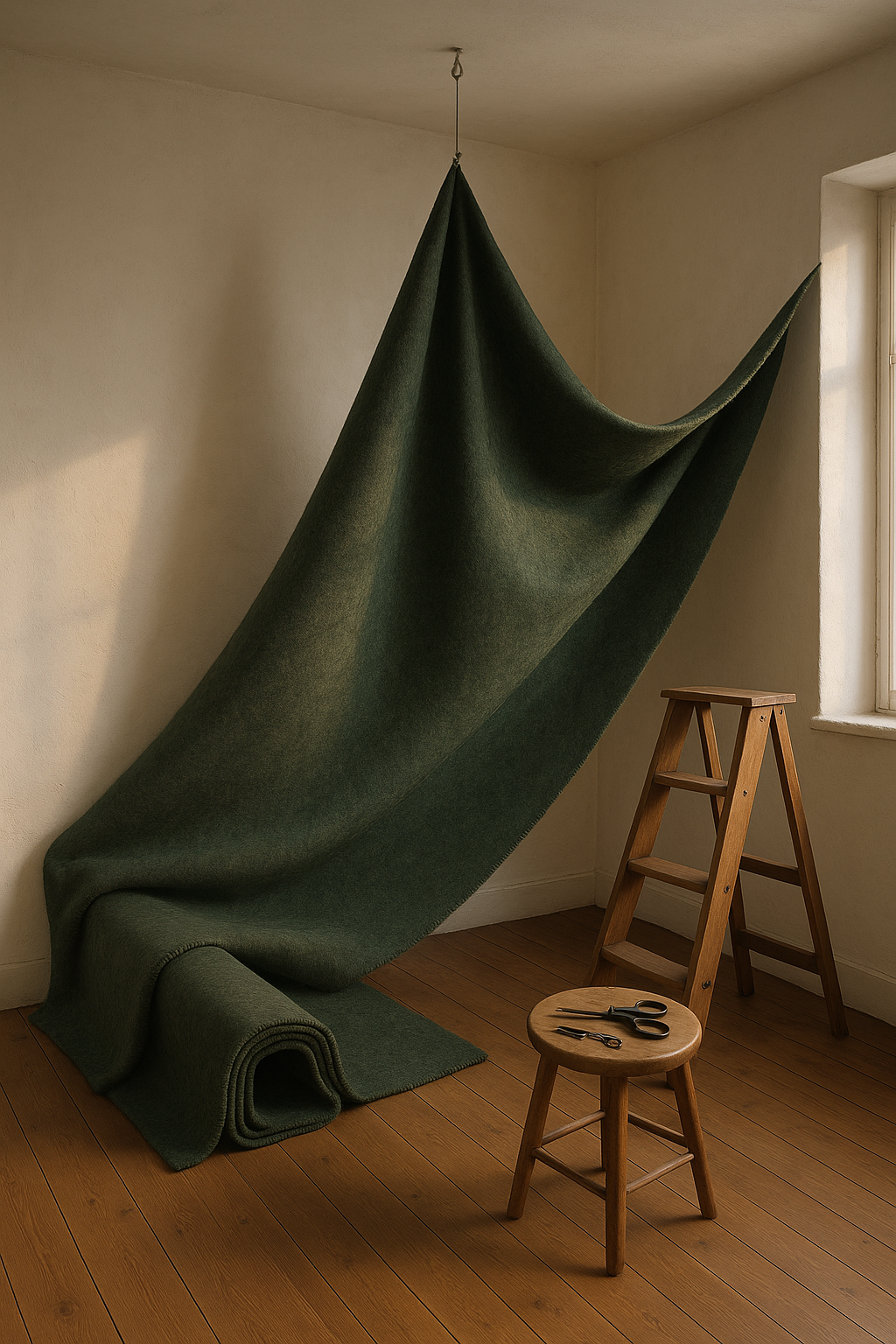}} & \adjustbox{valign=c}{\includegraphics[width=3.5cm, height=3.5cm]{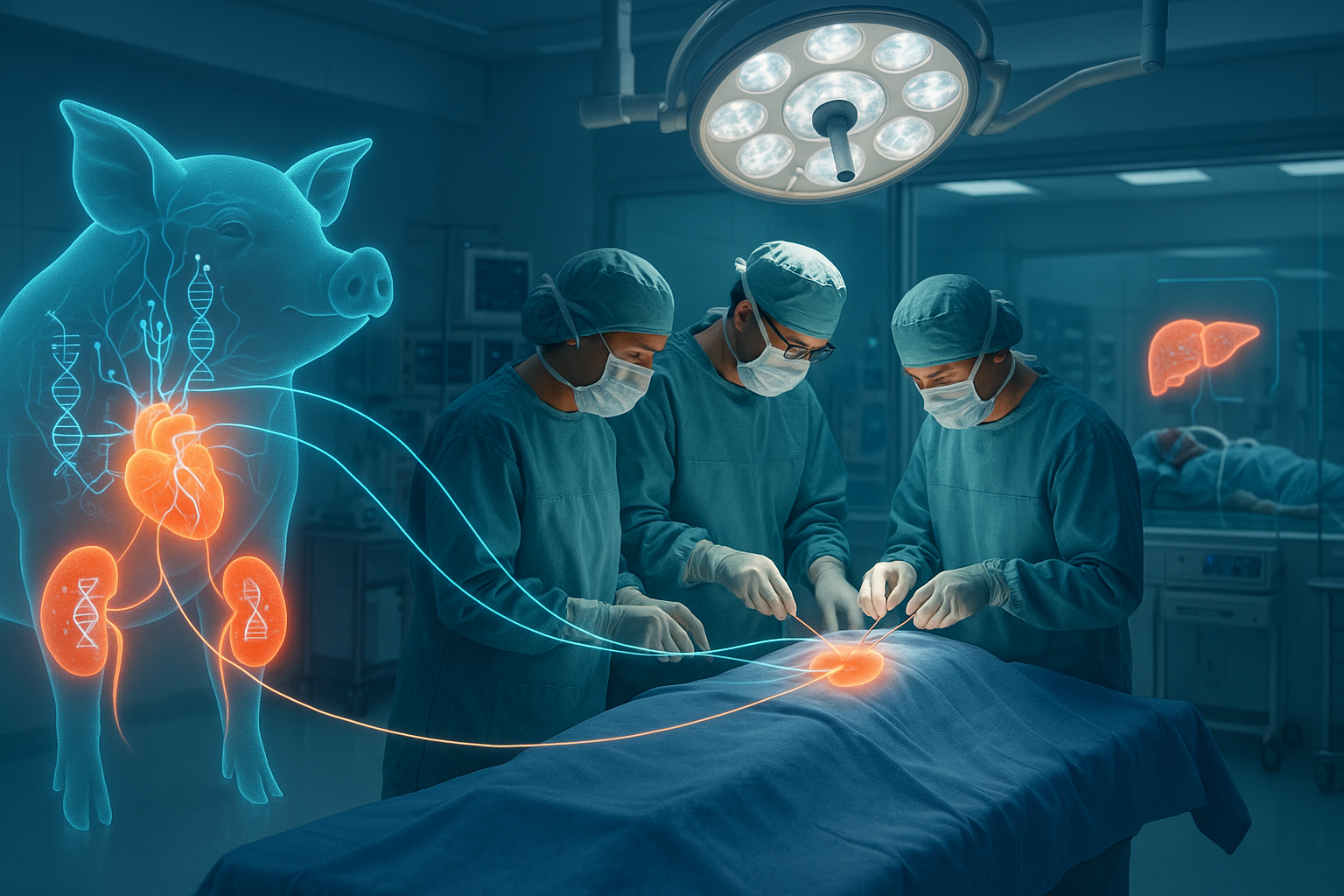}} \\
    \bottomrule
    \end{tabular}
    \caption{An example sentence with relevant and irrelevant image as context.}
    \label{tab:sent-image-example}
\end{table*}

We collected 75 English sentences between the length of 25 and 40 words, from news (the Guardian, CNN, Washington Post, Wallstreet Journal, BBC), books (Google Books) and Wikipedia. To minimize bias from data contamination, we selected sentences from 2025, with the exception of Wikipedia, for which it is hard to extract new sentences only.
We subjected these original English sentences to a round-trip translation using publicly available \emph{Moses}
models \citep{koehn-etal-2007-moses},\footnote{\url{https://www.statmt.org/moses-release/RELEASE-3.0/}} to introduce lexical, syntactic, and semantic infelicities to the sentences, following \citet{Lau&etal2020}. We used \emph{Moses} because most decoder only LLM-based MT systems generate fluent well-formed text. We employed three non-English languages as the pivot language (i.e., en$\rightarrow$cs$\rightarrow$en, en$\rightarrow$fr$\rightarrow$en, en$\rightarrow$de$\rightarrow$en) to create 75 $\times$ 3 round-trip translated sentences (see Table~\ref{tab:sentence-examples} for examples). We generated images for 75 original sentences, using GPT-5 with the prompt ``\textit{Generate an image that describes or is relevant to this sentence}". We inspected the generated images and observed that the quality was high in terms of their relevance to the original sentence (see Table~\ref{tab:sent-image-example}).\footnote{We acknowledge that the AI-generated images may contain hallucinations and imperfections. However, we opted for this approach --- as opposed to using existing image captioning dataset --- to mitigate the risk of data contamination. Additionally, generating the images allowed us to ensure tight congruence between the textual and visual modalities, so that the text is directly relevant to the image and vice versa.}
The total of 300 sentences (75 original, 225 modified) were split into multiple batches, so that each human participant would provide sentence acceptability ratings for 20 sentences (5 original, 15 modified) on a scale from 1 (very unnatural) to 4 (very natural).\footnote{Following \citet{Lau&etal2017}, we asked participants to evaluate \textit{naturalness} rather than \textit{acceptability}, a non-expert-friendly term.} The participants also indicated how concrete/abstract they found each sentence on a scale of 1 (very concrete) to 4 (very abstract). The sentences were presented in three different conditions (\textit{null}, \textit{relevant}, \textit{irrelevant}). In the \textit{null} condition, participants rated each sentence in terms of naturalness without any preceding visual context. In the \textit{relevant} and \textit{irrelevant} condition, participants rated each sentence after having seen a relevant or irrelevant visual context, respectively. The irrelevant images were paired with the sentences by a simplified permutation design (e.g., s\textsubscript{1}-im\textsubscript{2}, s\textsubscript{2}-im\textsubscript{3}, s\textsubscript{3}-im\textsubscript{1}).
To make sure that participants pay attention to the images, we asked them to select the most foregrounded object in the image from multiple choices, before rating the sentence. We batched the sentences such that no participant would see duplicate sentences or images. Each participant only rated sentences in one of three conditions.
We added two attention check questions, where a prompt to choose a specific answer was embedded in a usual sentence rating task.
Every sentence was rated in all three conditions (e.g., s\textsubscript{1}-\textit{null}, s\textsubscript{1}-\textit{relevant}, s\textsubscript{1}-\textit{irrelevant}). We recruited 20-25 native English-speaking participants for each condition and batch (gender-balanced). For quality control, we discarded data from participants who a) failed the attention check questions, b) rated original sentences as unnatural more than 40\% of the time, or c) selected incorrect answers to the image question more than 25\% of the time. As a result of such filtering criteria, 10\% of the total participants were removed from the collected data. Participants were 45 years old on average ($\pm$ 13), with 71\% British, 15\% American, and 14\% English speakers from other countries. The participants were paid GBP 9/hour.

\subsection{Results}

\begin{table}[h]
    \centering
    \begin{tabular}{cc@{\;\;}c@{\;\;}cc@{\;\;}c@{\;\;}c}
    \toprule 
         &  \multicolumn{3}{c}{\textbf{Original}} & \multicolumn{3}{c}{\textbf{Modified}}\\
         \midrule
         & N & R & I & N & R & I \\ 
         \midrule
    mean & 3.54 & 3.54 & 3.53 & 1.96 & 2.02 & 1.94 \\
    sd & 0.76 & 0.75 & 0.77 & 1.05 & 1.03 & 1.01  \\
    \bottomrule
    \end{tabular}
    \caption{Descriptive statistics for human ratings on sentence acceptability (1 - 4) in \textit{null} (N), \textit{relevant} (R), and \textit{irrelevant} (I) conditions.}
    \label{tab:human-descriptive}
\end{table}

\begin{figure*}
    \centering
    \includegraphics[width=1.1\linewidth]{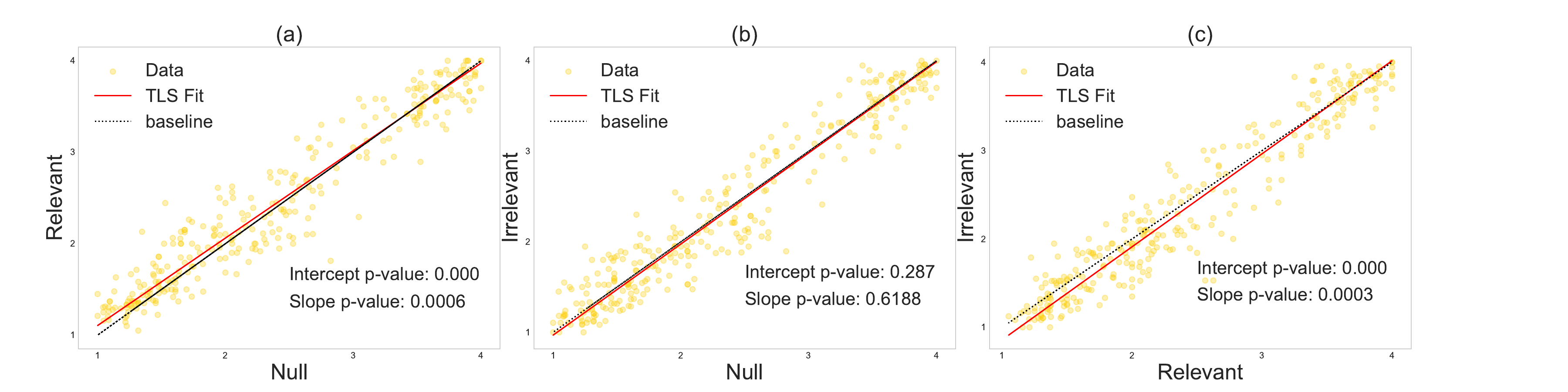}
    \caption{Human ratings of sentence acceptability in different conditions with regression lines generated from total least square regression. P-value alpha: 0.017 (Bonferroni correction).}
    \label{fig:humans}
    \includegraphics[width=1.1\linewidth]{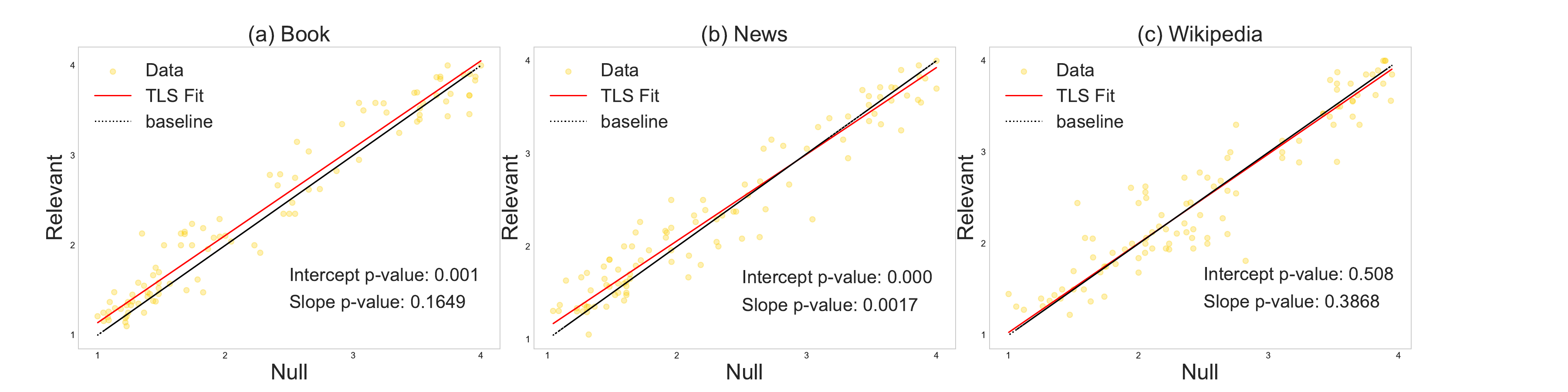}
    \caption{Human ratings of acceptability of sentences from different genres (books, news, Wikipedia) for \textit{null}-\textit{relevant} condition pair.}
    \label{fig:humans-genre-simple}
\end{figure*}

\begin{figure*}[t]
    \centering
    \begin{subfigure}[t]{0.32\linewidth}
        \centering
        \includegraphics[width=\linewidth]{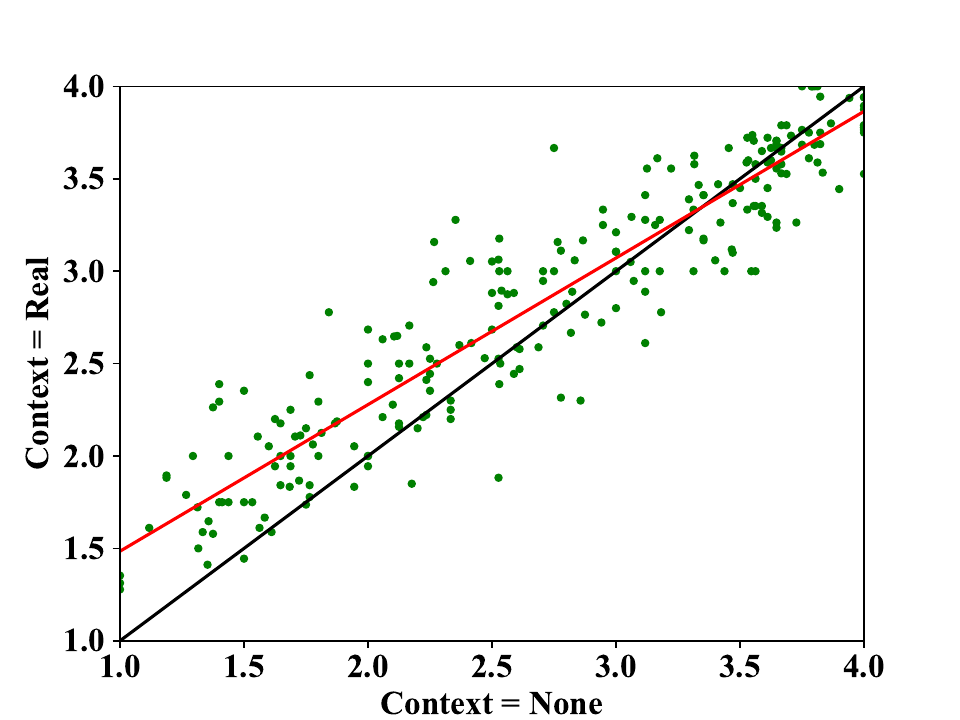}
        \caption{}
        \label{fig:main_a}
    \end{subfigure}\hfill
    \begin{subfigure}[t]{0.32\linewidth}
        \centering
        \includegraphics[width=\linewidth]{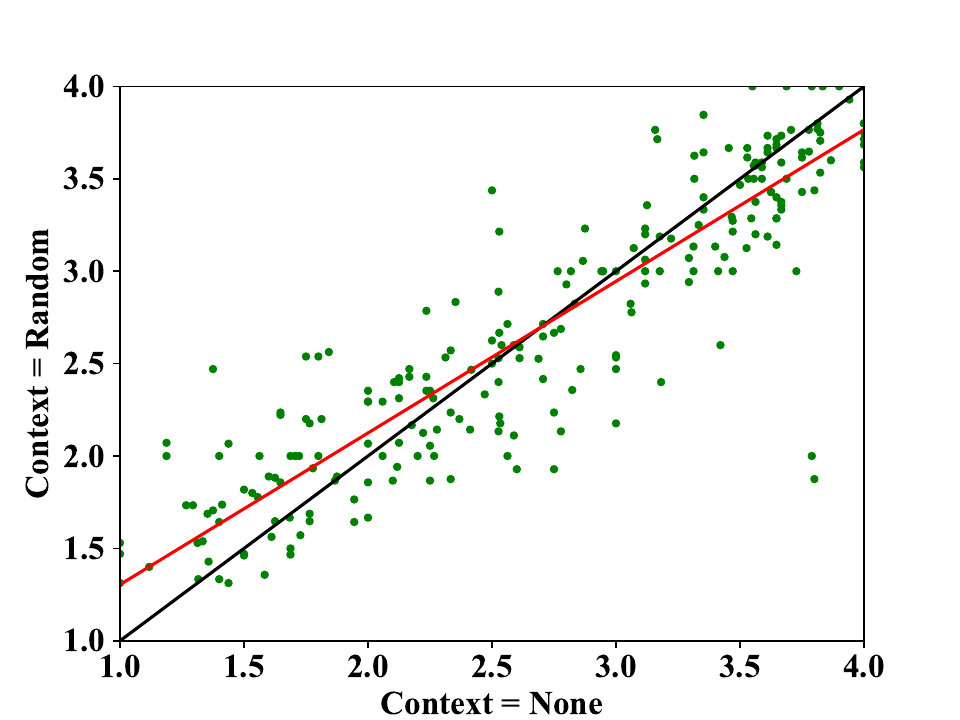}
        \caption{}
        \label{fig:main_b}
    \end{subfigure}\hfill
    \begin{subfigure}[t]{0.32\linewidth}
        \centering
        \includegraphics[width=\linewidth]{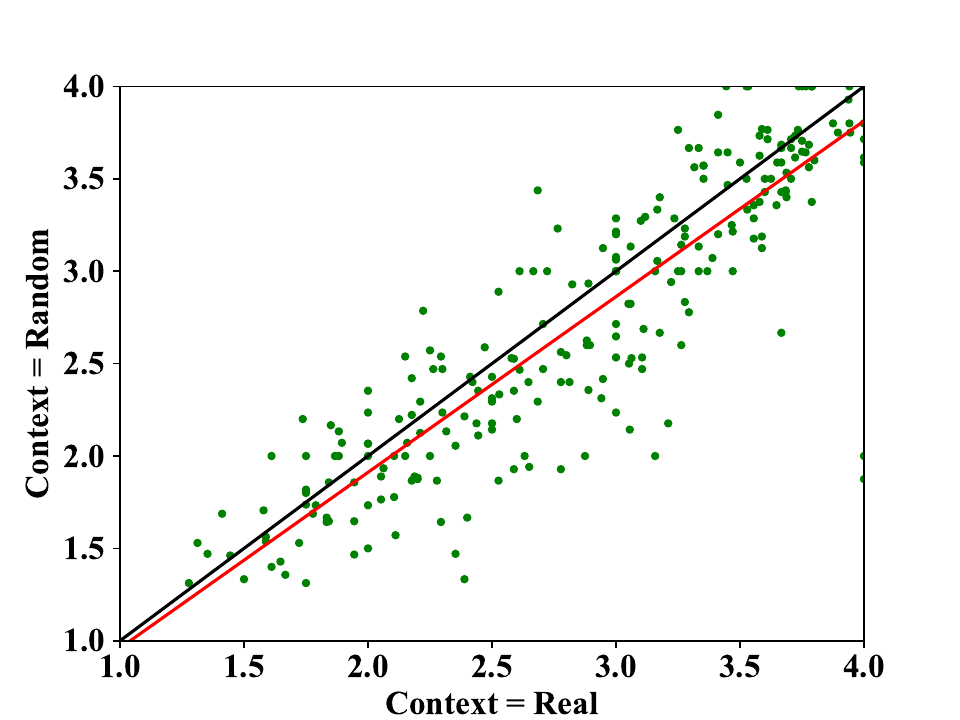}
        \caption{}
        \label{fig:main_c}
    \end{subfigure}
    \caption{Human ratings of sentence acceptability in different \textbf{textual} context. ``Context $=$ None'' means null context, ``Context $=$ Real'' means relevant context and ``Context $=$ Random'' means irrelevant context. Image reproduced from \citet{Lau&etal2020}.}
    \label{fig:lau-etal20}
\end{figure*}

Table~\ref{tab:human-descriptive} shows the average acceptability ratings for original and modified sentences. The average ratings for original sentences are higher than for modified sentences, showing that modified sentences are infelicitous, as expected. We calculated mean sentence acceptability scores for each sentence across multiple human raters in each of the three conditions. 
Figure~\ref{fig:humans} shows the average human ratings for each sentence in different conditions. Each data point (yellow dot)
represents the acceptability rating for a sentence. The red line in each pane represents the total least square regression fit for the data points in two out of three conditions at a time (\textit{null-relevant}, \textit{null-irrelevant}, \textit{relevant-irrelevant}). In (a), relative to the \textit{null} condition, we observe a slight raising effect in the \textit{relevant} condition, where sentence acceptability scores in the lower end are higher. This indicates that human raters tend to judge a bad sentence slightly less bad when presented with relevant visual context prior to the judgment. We observe no such effect for \textit{irrelevant} contexts (b), with the ratings for a sentence being almost identical in both conditions. The comparison between relevant and irrelevant in (c) confirms this, as the ratings for unnatural sentences are higher in relevant conditions, which mirrors (a). 

Our results do not generally exhibit the compression effect for humans, reported in \citet{Lau&etal2020} (Figure \ref{fig:lau-etal20}), where sentence acceptability ratings on both ends were compressed, causing bad sentences to be judged less bad, and good sentences to be judged less good, when a textual context precedes the target sentence.\footnote{\citet{Lappin2021} describes a total least square regression test which shows that the compression effect that \citet{Lau&etal2020} report is actual, and it cannot be reduced to regression to the norm.} In our results, we observe a raising effect only for the lower end (bad sentences) when relevant visual contexts are presented (see Figure~\ref{fig:humans}(a)). This would seem to indicate that a discourse coherence effect is operative in visual contexts for human processing, as it works only to (slightly) raise lower end judgments when the preceding image renders the text more accessible. On the discourse coherence account, infelicitous sentences look more natural when placed within a context that makes them appear more coherent (see \S\ref{sec:introduction} for discussion regarding the cognitive load and discourse coherence accounts of compression).

Genre is also relevant. One possibility is that the characteristics of the 
sentences we tested, represented by the genre (news, books, Wikipedia) produces different effects for preceding contexts on human sentence processing. We looked
at the effect of genre of sentences. Figure~\ref{fig:humans-genre-simple} shows the same information about human sentence acceptability judgment, on sentences from books, news, and Wikipedia, respectively. For sentences sourced from books, we observe a more uniform raising effect, while for sentences from news we observe a compression effect, though the magnitude is much smaller than the one reported in \citet{Lau&etal2020}. Visual contexts for sentences of distinct genres may affect acceptability judgmens differently. Sentences from books are rated as more abstract than sentences from news or wikipedia (see Table~\ref{tab:human-abstract}). Sentences from books are rated as more natural when relevant visual contexts are presented prior to the sentences, because images may render the content of ill-formed sentences more accessible. Sentences that are more concrete show a pattern comparable to the compression effect, which was manifest in textual contexts, but this effect is minimal.  We do not observe it in irrelevant visual contexts, unlike \citet{Lau&etal2020}, where it is apparent in both non-null textual environments. 

\begin{table}[h]
    \centering
    \begin{tabular}{cccc}
    \toprule 
    & \textbf{Books} & \textbf{News} & \textbf{Wikipedia} \\ 
    \midrule
    mean & 1.95 & 1.63 & 1.70  \\
    sd & 1.03 & 0.86 & 0.92 \\
    \bottomrule
    \end{tabular}
    \caption{Mean abstractness scores reported by human participants on original English sentences from books, news, and Wikipedia.}
    \label{tab:human-abstract}
\end{table}

Another possible factor conditioning the absence of the compression effect involves cognitive load. According to this account, interpreting textual context puts additional informational processing demands on humans, forcing the acceptability ratings to be more conservative, clustering towards the middle range. 
This mechanism does not appear to operate in the case of visual images in conjunction with sentences. It is easier for humans to ignore them if they find no connection between an image and a sentence. Ignoring \textit{textual} context, relevant or irrelevant, is presumably more difficult, as it is interpreted through the same processing mode. This may explain why humans exhibit a clear compression effect only with textual contexts. As a general note, the characteristics of visual contexts and textual contexts may 
differ in informational role. A visual context can be a depiction of a sentence's content, while a textual context is a preceding set of sentences that provides a prologue to the target sentence. 

\section{LLM Acceptability Judgments}
\label{sec:models}

We conducted two types of analyses with LLMs.
First, we prompted LLMs (\S\ref{sec:model-exp1}) to provide sentence acceptability judgments in the identical setup as the human experiment in \S\ref{sec:humans}. We then looked at the internal representations of the open-source LLMs by calculating normalised logprob values (\S\ref{sec:model-exp2}), based on token probabilities, following prior work \citep{Lau&etal2020, kauf-etal-2024-log}. We used five open-source models that can take image and text as input (vision and language models) -- \intvls, \intvlb~\cite{internvl3-2024}, \qwens, \qwenb~\citep{qwen2.5-VL}, and \llava~\cite{llava2024} -- and two closed models -- \gpt \& \gptmini~\cite{openai2024gpt4technicalreport}. \intvls,~\intvlb, have 1B and 8B parameters, respectively. \qwens,~\qwenb~have 3B and 7B, respectively, and \llava~has 7B parameters.  We also made sure to select models released in early 2025 at the latest, to minimize the possibility of data contamination, although it should be noted that the chance of contamination is high for sentences from Wikipedia. We did our probability-based experiment (\S\ref{sec:model-exp2}) on open-source models, because closed models do not allow for the extraction of logprob values for their output.

\subsection{Prompting-based analysis}
\label{sec:model-exp1}
We prompted the LLMs in a zero-shot setting to judge the naturalness of each of the 300 sentences used in \S\ref{sec:humans} (75 original sentences + 75 $\times$ 3 modified sentences), under three different conditions. In the \textit{null} condition, we presented only the sentences to the LLMs. In the \textit{relevant} and \textit{irrelevant} conditions, we first prompted the LLMs to see an image and solve the same task given to the human participants in \S\ref{sec:humans}. We then prompted them to judge the naturalness of the subsequent sentence after feeding their response to the next generation call (i.e., in the second call the input concatenates the image attention check task and response, and the sentence acceptability task instruction). We did not provide any explicit instructions to connect the image with the following sentence. We used 10 initialization seeds for each model, and we averaged the sentence acceptability ratings across the seeds. We used hyper-parameter settings of temperature=0.7, top-p=1.0, top-k=50 for all open-source models, and the default settings for closed models.

\begin{table}[h]
    \centering
    \resizebox{\linewidth}{!}{\begin{tabular}{cccc@{\;\;}c@{\;\;}c}
    \toprule 
    & & \textbf{All} & \textbf{N} & \textbf{R} & \textbf{I} \\
    \midrule
   \multirow{7}{*}{\makecell[c]{LLM ratings $\sim$ \\ Human ratings}} & \gpt  & 0.87& \textbf{0.89} &0.87 & 0.88  \\
    & \gptmini  & 0.89&\textbf{0.89} &\textbf{0.89} & 0.88  \\
    \cmidrule{2-6}
 & \intvls  & 0.32&\textbf{0.66} & 0.26 & 0.16 \\
    & \intvlb  & 0.83& \textbf{0.88} &0.85 & 0.85\\
    & \qwens   & 0.65& \textbf{0.70} & 0.61 & 0.66  \\
    & \qwenb  & 0.78& \textbf{0.84}&0.75 & 0.76  \\
    & \llava  & 0.25 &\textbf{0.39} &0.21 & 0.15  \\
    \bottomrule
    \end{tabular}}
    \caption{Spearman ($\rho$) correlations between average human sentence acceptability ratings (1-4) and LLM-prompted ratings (1-4) in null (N), relevant (R) and irrelevant (I) conditions. All correlations significant (p$<$0.001). Highest correlations per row marked in bold.}
    \label{tab:corr-human-model-ratings}
\end{table}

\subsection{Probability-based analysis} 
\label{sec:model-exp2}
For this analysis we used the sentence probability as estimated by the LLMs directly \citep{ide-etal-2025-make, hu-2024, kauf-etal-2024-log}. We passed the images and sentences through the open-source models (forward pass without generation) and extracted logits for only the input sentence in all three conditions (\textit{null}, \textit{relevant}, \textit{irrelevant}). We calculated the average log probabilities of a sentence, normalised by the number of tokens (MeanLP), by summing the log-probabilities of each token in a sentence, conditioned on the preceding tokens \cite{Lau&etal2020, kauf-etal-2024-log}. We considered the MeanLP as a proxy for the LLM's internal representation of its sentence acceptability judgment, following prior work. We experimented with normalising functions employed in \citet{Lau&etal2020}, and found them to be unnecessary based on the high correlations of most LLMs with the human ratings.

\begin{table}[h]
    \centering
   \resizebox{\columnwidth}{!}{
   \begin{tabular}{cccc@{\;\;}c@{\;\;}c}
    \toprule 
    & & \textbf{All} & \textbf{N} & \textbf{R} & \textbf{I} \\
    \midrule
    \multirow{5}{*}{\makecell[c]{MeanLP $\sim$ \\ Human ratings}} & \intvls &  0.70 & \textbf{0.72} & 0.70 & 0.70 \\
    & \intvlb   & 0.70 & \textbf{0.72 }& 0.71 & 0.70\\
    & \qwens   & 0.61 & \textbf{0.70} & 0.61 & 0.57   \\
    & \qwenb & 0.73 & \textbf{0.79} & 0.76 & 0.71 \\
    & \llava & 0.74 & 0.74 & \textbf{0.76} & 0.73
    \\
    \bottomrule
    \end{tabular}}
    \caption{Spearman ($\rho$) correlations between Mean Logprobs from open-source LLMs and average human sentence acceptability ratings. All correlations significant (p$<$0.001). Highest correlations per row marked in bold.}
    \label{tab:corr-human-model-prob}
\end{table}

\subsection{Results} 

Table~\ref{tab:corr-human-model-ratings} shows correlations (Spearman  $\rho$) between average human sentence acceptability ratings and LLM-generated ratings (\S\ref{sec:model-exp1}).\footnote{We only report Spearman $\rho$, as Pearson r yielded very similar results. The same applies to Tables~\ref{tab:corr-human-model-prob} and \ref{tab:corr-model-rating-model-prob}.} The two GPT models show overall correlations of over 0.87, with little to no variations across conditions. Of the open-source LLMs, models with higher parameter size (7B-8B) perform almost equally well, with the exception of \llava. The poor performance of the smaller models (\intvls, \qwens) and \llava might be attributed to their reduced capability to follow instructions (\llava in particular is not trained to follow instructions). The open-source LLMs also seem to perform better without any visual context (N).

\begin{figure*}[h!]
    \centering
    \includegraphics[width=1.1\linewidth]{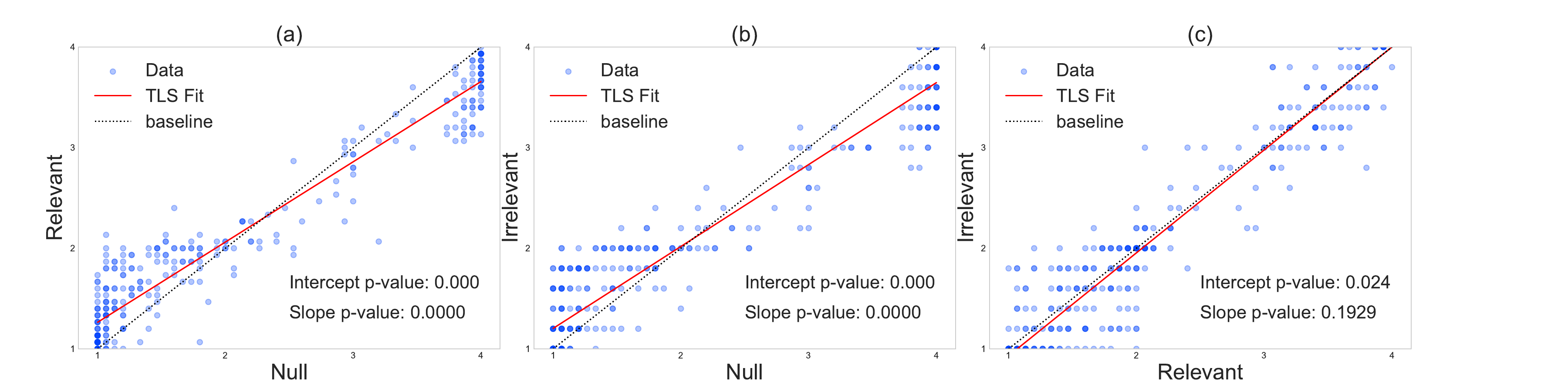}
    \caption{Sentence acceptability ratings generated by \gpt.}
    \label{fig:ratings-gpt4o}
\end{figure*}

\begin{figure*}[h!]
    \centering
    \includegraphics[width=1.1\linewidth]{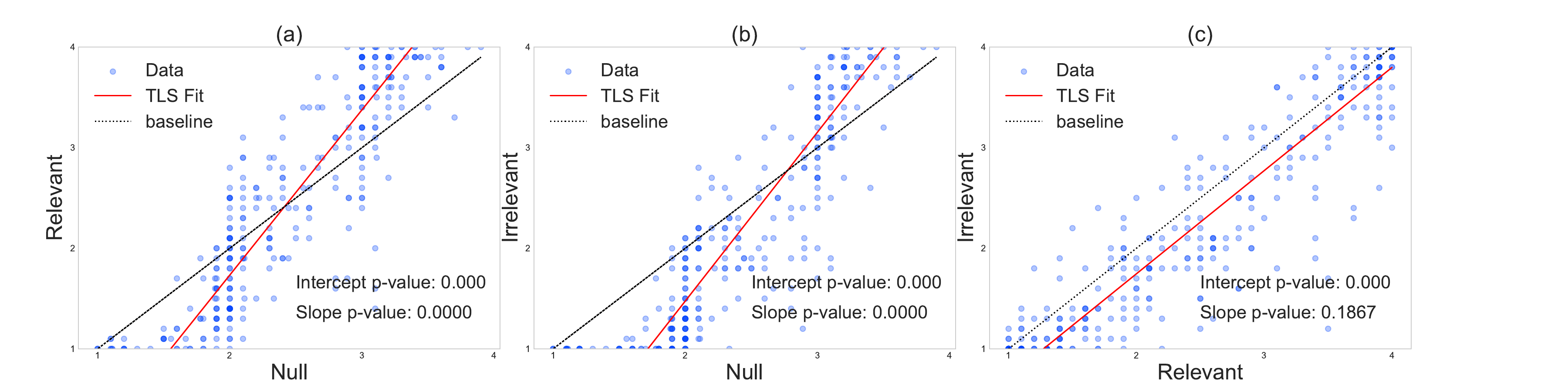}
    \caption{Sentence acceptability ratings generated by \intvlb.}
    \label{fig:ratings-internvlbig}
\end{figure*}

\begin{figure*}[h!]
    \centering
    \includegraphics[width=1.1\linewidth]{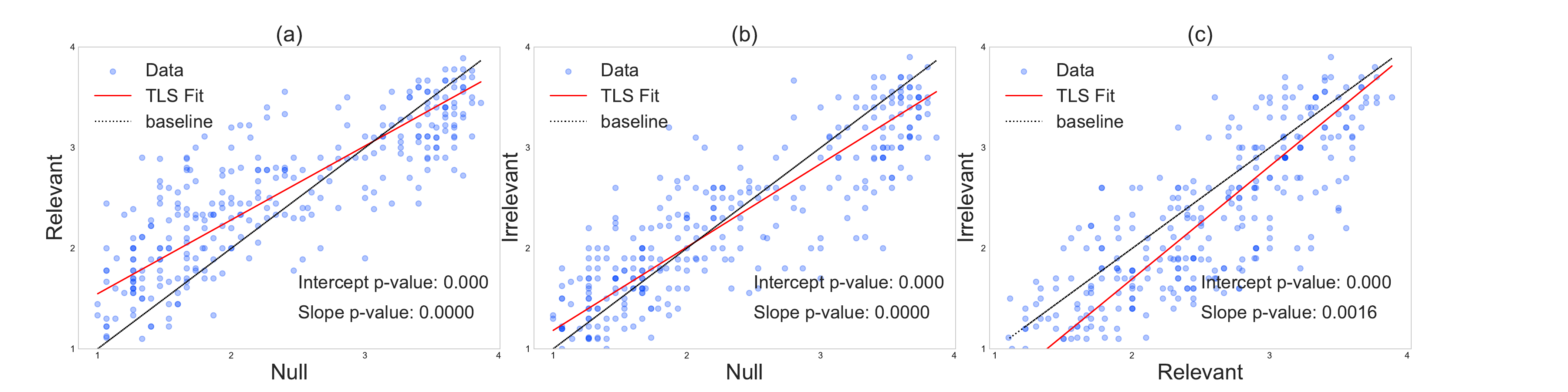}
    \caption{Sentence acceptability ratings generated by \qwenb.}
    \label{fig:ratings-qwenbig}
\end{figure*}

Table~\ref{tab:corr-human-model-prob} presents correlations between average human sentence acceptability ratings and probability measures from the LLMs described in \S\ref{sec:model-exp2}. Compared to the prompting results (Table~\ref{tab:corr-human-model-ratings}), MeanLP generally performs on par or better for the open-sourced LLMs. Most interestingly, the performance of these models \textit{across conditions} (N, R, I) also appears to be much more consistent --- a contrast compared to the prompting results in the case of \llava. 

To better understand the similarity between MeanLP versus prompted model ratings, we present their correlation in Table~\ref{tab:corr-model-rating-model-prob}. The larger models (\intvlb and \qwenb) have better correlations (both 0.70 for \textit{All}), while the others have much lower correlations. Either way, these results show that these two approaches produce somewhat distinct predictions of acceptability ratings. But, Tables~\ref{tab:corr-human-model-ratings}, \ref{tab:corr-human-model-prob}, and \ref{tab:corr-model-rating-model-prob} all suggest that correlations are higher in the \textit{null} condition compared to the non-\textit{null} conditions (with the exception of one case involving \llava).

\begin{table}[h]
    \centering
   \resizebox{\columnwidth}{!}{
   \begin{tabular}{cccc@{\;\;}c@{\;\;}c}
    \toprule 
    & & \textbf{All} & \textbf{N} & \textbf{R} & \textbf{I} \\
    \midrule
    \multirow{5}{*}{\makecell[c]{MeanLP $\sim$ \\ Model ratings}} & \intvls
    &  0.30 &\textbf{0.58}&0.30&0.17\\
    & \intvlb
    &  0.70&\textbf{0.75}&0.74&0.71\\
    & \qwens
    &   0.52&\textbf{0.56}&0.48&0.45  \\
    & \qwenb
    &  0.70&\textbf{0.76}&0.71&0.68 \\
    & \llava
    &   0.28&\textbf{0.35}&0.24&0.22
    \\
    \bottomrule
    \end{tabular}}
    \caption{Spearman ($\rho$) correlations between Mean Logprobs from open-source LLMs and model prompted ratings. All correlations significant (p$<$0.001). Highest correlations per row marked in bold.}
    \label{tab:corr-model-rating-model-prob}
\end{table}
Based on the correlation analyses, we consider \qwenb, \intvlb, and \gpt~to have yielded the most reliable results. In Figures~\ref{fig:ratings-gpt4o}, \ref{fig:ratings-internvlbig}, and \ref{fig:ratings-qwenbig} we present the scatterplots of the sentence acceptability ratings produced by these models in each condition pair (\textit{null-relevant}, \textit{null-irrelevant}, \textit{relevant-irrelevant}) for more detail, parallel to the human results in \S\ref{sec:humans}. Our intention is to examine these scatterplots to understand the qualitative nature of the model rating distribution, and also to make sure the correlation numbers are not skewed by a few outliers.
In general, we see that the distributions of sentence acceptability ratings produced by the LLMs appear quite different from those of humans. \gpt~-generated ratings cluster around top and bottom ends, and \intvlb~-generated ratings cluster around 2 and 3.
\qwenb~shows distributions that are most similar to humans,\footnote{This might be an artifact of \qwenb~producing more varied ratings with different initialisation seeds than the other LLMs.} but with a much higher variance than humans. Interestingly, we observe the compression effect for both \gpt~and \qwenb. \citet{Lau&etal2020} show a very similar compression effect for human sentence ratings in textual contexts, as we see in Figure~\ref{fig:lau-etal20}. But we observe the opposite of compression for \intvlb. We do not have an explanation for this contrast. We will be exploring it in future work. The general divergence in the distributions of model-generated ratings from human-generated ratings may be due to the different mechanisms by which LLMs process sentences in comparison to humans. LLMs have less memory interference, and they retain perfect access to previous words \cite{Oh&Linzen2025}.

\section{Discussion}

The work that we present here indicates that LLMs achieve a high degree of accuracy in predicting human sentence acceptability judgments. It also shows that the normalised logprob values that these models assign to sentences are a reliable predictor of human ratings for sentence acceptability. Despite the strong convergence of LLM logprob scoring and human naturalness rating, most of the LLMs that we consider in this work display different data clustering patterns than humans in the distributions of these judgments (cf. Figures~\ref{fig:humans} and \ref{fig:ratings-internvlbig} for two illustrative examples). In particular, \gpt~exhibits a more polarised pattern, while \qwenb~approaches human distributions (Figure \ref{fig:ratings-gpt4o} vs \ref{fig:ratings-qwenbig}). However, although \qwenb~most closely approximates the human distributions, it still differs from them, both in our experiments and in those reported by \citet{Lau&etal2020} (cf. Figures~\ref{fig:humans}, \ref{fig:lau-etal20}, and \ref{fig:ratings-qwenbig}). Due to their greater memory capacity compared to humans, \citet{Oh&Linzen2025} argue that LLMs' superhuman next-token prediction ability makes them unsuitable as cognitive models of human linguistic prediction. However, the tasks differ. \citet{Oh&Linzen2025} focus on linguistic continuation tasks which effectively involve next-token prediction. In contrast, our experiments examine acceptability judgments, where LLMs predict evaluation scores and not upcoming tokens. It therefore remains an open question whether the differing patterns we observe reflect the LLMs' superhuman predictive capacity or instead stem from fundamentally different processing strategies. 

Our experiments, viewed from the perspective of the work reported in \citet{Bernardy&etal2018,Lau&etal2020}, suggest an interesting difference in the way that humans and LLMs process information in different modalities. Textual context, whether relevant or not to the following sentence, influences the way in which humans assess the naturalness of sentences, both well-formed ones, and those containing infelicities. Interpreting these contexts requires processing resources that influence judgments concerning the naturalness of following sentences. 

By contrast, for humans, interpreting images seems to proceed through alternative processing mechanisms, which permit the image to be discarded, or suppressed, when rating the naturalness of a following sentence. This effect is particularly clear when the image has no clear relation to the sentence. However, LLMs, at least the vision and language models
we used in this work, incorporate images into their sentence processing environment. They include the images when assigning probability values to sentences, and in rating them for naturalness.
They do not discard them, even when they are irrelevant. Cognitive load appears to  affect their judgments across visual and textual modalities. This is not surprising since their architectures are generally designed to maximise contexts \citep{Oh&Linzen2025}. The vectors that encode these contexts include elements of all modalities that the model attends to.\footnote{Our experimental results are broadly consistent with the view of the relation between LLMs and human processing presented in \citet{Oh&Linzen2025}. Simlarly, our suggestions for future work align well with their approach.}

\section{Future Work}

In future work we will examine more closely the role  of different types of text genre in determining the impact of both textual and visual contexts on human sentence acceptability rating. We will also explore the mechanisms through which humans suppress images, but not preceding text, when processing sentences. Our objective here is to obtain a clear sense of the way in which cognitive load is conditioned by different sorts of context, relative to distinct text genres. For this, we will also consider other linguistic tasks than sentence acceptability judgment for better generalizations.

Another issue that we will explore is how LLMs can be modified to converge on human processing with respect to the suppression of images in the assignment of logprob values to sentences, without losing the content of these images. This would cause the modified model to more closely approximate observed human sentence processing in multimodal environments. 

\section*{Acknowledgements}
The work reported in this paper was supported by a grant from the Swedish Research Council (VR project 2014-39) for the establishment of the Centre for Linguistic Theory and Studies in Probability (CLASP) at the University of Gothenburg. The computation and data storage for our experiments were supported by the National Academic Infrastructure for Supercomputing in Sweden (NAISS), partially funded by the Swedish Research Council through grant agreement no. 2022-06725.

\section*{Limitations}

The work reported in this paper only addresses English. We tried to minimise data contamination issues by experimenting with sentences from 2025, but there is a high chance of data contamination with  sentences from Wikipedia, which are not as new. We reported that genre effects influence sentence acceptability judgment by humans, but the explored genres in this work are limited to three (news, books, wikipedia). We did not directly compare visual contexts and document contexts for the same target sentence, which may render the results inconclusive. Lastly, we did not fully address the characteristic differences of visual versus textual contexts in this work. We will follow up on these limitations in our future work.

\section*{Ethics statement}

All our human participants took part in the experiments voluntarily, and received due compensations. We recognized that we might expose crowdworkers to disturbing/offensive images and texts. To minimize this risk, we manually examined the 75 sentences before administering the human experiments. GPT-5 also has its own guardrails for image generation to keep such risks low (\url{https://cdn.openai.com/gpt-5-system-card.pdf}). Some of our modeling experiments were done with closed models, which may render the results not entirely transparent. We did not rely on any AI-assistant tools for the generation of our article text. 

\bibliographystyle{lrec2026-natbib}
\bibliography{lrec2026-example}

@InProceedings{Bernardy&etal2018,
  author    = {Bernardy, Jean-Philippe and Lappin, Shalom and Lau, Jey Han},
  title     = {The Influence of Context on Sentence Acceptability Judgements},
  booktitle = {Proceedings of the 56th Annual Meeting of the Association for 
               Computational Linguistics (ACL 2018)},
  year      = {2018},
  address   = {Melbourne, Australia},
  pages     = "456--461",
}

@book{Lappin2021,
	Address = {Boca Raton, London, New York},
	Author = {Shalom Lappin},
	Publisher = {CRC Press, Taylor \& Francis},
	Title = {Deep Learning and Linguistic Representation},
	Year = {2021},
}

@article{Lau&etal2017,
	title = {Grammaticality, {Acceptability}, and {Probability}: {A} {Probabilistic} {View} of {Linguistic} {Knowledge}},
	volume = {41},
	journal = {Cognitive Science},
	author = {Lau, Jey Han and Clark, Alexander and Lappin, Shalom},
	year = {2017},
	pages = {1202--1241}
}

@article{Lau&etal2020,
author = {Lau, Jey Han and Armendariz, Carlos and Lappin, Shalom and Purver, Matthew and Shu, Chang},
title = {How Furiously Can Colorless Green Ideas Sleep? Sentence Acceptability in Context},
journal = {Transactions of the Association for Computational Linguistics},
volume = {8},
number = {},
pages = {296--310},
year = {2020},
doi = {10.1162/tacl\_a\_00315},
}

@article{warstadt-etal-2019-neural,
    title = "Neural Network Acceptability Judgments",
    author = "Warstadt, Alex  and
      Singh, Amanpreet  and
      Bowman, Samuel R.",
    editor = "Lee, Lillian  and
      Johnson, Mark  and
      Roark, Brian  and
      Nenkova, Ani",
    journal = "Transactions of the Association for Computational Linguistics",
    volume = "7",
    year = "2019",
    address = "Cambridge, MA",
    publisher = "MIT Press",
    url = "https://aclanthology.org/Q19-1040/",
    doi = "10.1162/tacl_a_00290",
    pages = "625--641",
}

@inproceedings{qiu&etal2024,
    title = "Evaluating Grammatical Well-Formedness in Large Language Models: A 
             Comparative Study with Human Judgments",
    author = "Qiu, Zhuang  and Duan, Xufeng  and Cai, Zhenguang",
    editor = "Kuribayashi, Tatsuki  and Rambelli, Giulia  and Takmaz, Ece  and
              Wicke, Philipp  and Oseki, Yohei",
    booktitle = "Proceedings of the Workshop on Cognitive Modeling and Computational 
                 Linguistics",
    month = Aug,
    year = "2024",
    address = "Bangkok, Thailand",
    publisher = "Association for Computational Linguistics",
    doi = "10.18653/v1/2024.cmcl-1.16",
    pages = "189--198",
}

@article{hu-2024,
author = {Jennifer Hu  and Kyle Mahowald  and Gary Lupyan  and Anna Ivanova  and Roger Levy },
title = {Language models align with human judgments on key grammatical constructions},
journal = {Proceedings of the National Academy of Sciences},
volume = {121},
number = {36},
pages = {e2400917121},
year = {2024},
doi = {10.1073/pnas.2400917121},}

@inproceedings{ide-etal-2025-make,
    title = "How to Make the Most of {LLM}s' Grammatical Knowledge for Acceptability Judgments",
    author = "Ide, Yusuke  and
      Nishida, Yuto  and
      Vasselli, Justin  and
      Oba, Miyu  and
      Sakai, Yusuke  and
      Kamigaito, Hidetaka  and
      Watanabe, Taro",
    editor = "Chiruzzo, Luis  and
      Ritter, Alan  and
      Wang, Lu",
    booktitle = "Proceedings of the 2025 Conference of the Nations of the Americas Chapter of the Association for Computational Linguistics: Human Language Technologies (Volume 1: Long Papers)",
    month = apr,
    year = "2025",
    address = "Albuquerque, New Mexico",
    publisher = "Association for Computational Linguistics",
    url = "https://aclanthology.org/2025.naacl-long.380/",
    doi = "10.18653/v1/2025.naacl-long.380",
    pages = "7416--7432",
    ISBN = "979-8-89176-189-6",
}

@inproceedings{kauf-etal-2024-log,
    title = "Log Probabilities Are a Reliable Estimate of Semantic Plausibility in Base and Instruction-Tuned Language Models",
    author = "Kauf, Carina  and
      Chersoni, Emmanuele  and
      Lenci, Alessandro  and
      Fedorenko, Evelina  and
      Ivanova, Anna A",
    editor = "Belinkov, Yonatan  and
      Kim, Najoung  and
      Jumelet, Jaap  and
      Mohebbi, Hosein  and
      Mueller, Aaron  and
      Chen, Hanjie",
    booktitle = "Proceedings of the 7th BlackboxNLP Workshop: Analyzing and Interpreting Neural Networks for NLP",
    month = nov,
    year = "2024",
    address = "Miami, Florida, US",
    publisher = "Association for Computational Linguistics",
    url = "https://aclanthology.org/2024.blackboxnlp-1.18/",
    doi = "10.18653/v1/2024.blackboxnlp-1.18",
    pages = "263--277",
}

@article{llava2024,
      title={Improved Baselines with Visual Instruction Tuning}, 
      author={Haotian Liu and Chunyuan Li and Yuheng Li and Yong Jae Lee},
      year={2024},
      journal={ar{X}iv preprint arXiv:2310.03744},
      eprint={2310.03744},
      primaryClass={cs.CV},
      url={https://arxiv.org/abs/2310.03744}, 
}

@inproceedings{koehn-etal-2007-moses,
    title = "{M}oses: Open Source Toolkit for Statistical Machine Translation",
    author = "Koehn, Philipp  and
      Hoang, Hieu  and
      Birch, Alexandra  and
      Callison-Burch, Chris  and
      Federico, Marcello  and
      Bertoldi, Nicola  and
      Cowan, Brooke  and
      Shen, Wade  and
      Moran, Christine  and
      Zens, Richard  and
      Dyer, Chris  and
      Bojar, Ond{\v{r}}ej  and
      Constantin, Alexandra  and
      Herbst, Evan",
    editor = "Ananiadou, Sophia",
    booktitle = "Proceedings of the 45th Annual Meeting of the Association for Computational Linguistics Companion Volume Proceedings of the Demo and Poster Sessions",
    month = jun,
    year = "2007",
    address = "Prague, Czech Republic",
    publisher = "Association for Computational Linguistics",
    url = "https://aclanthology.org/P07-2045/",
    pages = "177--180"
}

@misc{qwen2.5-VL,
    title = {Qwen2.5-VL},
    url = {https://qwenlm.github.io/blog/qwen2.5-vl/},
    author = {Qwen Team},
    month = {January},
    year = {2025}
}

@misc{openai2024gpt4technicalreport,
      title={GPT-4 Technical Report}, 
      author={OpenAI and Josh Achiam and Steven Adler and Sandhini Agarwal and Lama Ahmad and Ilge Akkaya and Florencia Leoni Aleman and Diogo Almeida and Janko Altenschmidt and Sam Altman and Shyamal Anadkat and Red Avila and Igor Babuschkin and Suchir Balaji and Valerie Balcom and Paul Baltescu and Haiming Bao and Mohammad Bavarian and Jeff Belgum and Irwan Bello and Jake Berdine and Gabriel Bernadett-Shapiro and Christopher Berner and Lenny Bogdonoff and Oleg Boiko and Madelaine Boyd and Anna-Luisa Brakman and Greg Brockman and Tim Brooks and Miles Brundage and Kevin Button and Trevor Cai and Rosie Campbell and Andrew Cann and Brittany Carey and Chelsea Carlson and Rory Carmichael and Brooke Chan and Che Chang and Fotis Chantzis and Derek Chen and Sully Chen and Ruby Chen and Jason Chen and Mark Chen and Ben Chess and Chester Cho and Casey Chu and Hyung Won Chung and Dave Cummings and Jeremiah Currier and Yunxing Dai and Cory Decareaux and Thomas Degry and Noah Deutsch and Damien Deville and Arka Dhar and David Dohan and Steve Dowling and Sheila Dunning and Adrien Ecoffet and Atty Eleti and Tyna Eloundou and David Farhi and Liam Fedus and Niko Felix and Simón Posada Fishman and Juston Forte and Isabella Fulford and Leo Gao and Elie Georges and Christian Gibson and Vik Goel and Tarun Gogineni and Gabriel Goh and Rapha Gontijo-Lopes and Jonathan Gordon and Morgan Grafstein and Scott Gray and Ryan Greene and Joshua Gross and Shixiang Shane Gu and Yufei Guo and Chris Hallacy and Jesse Han and Jeff Harris and Yuchen He and Mike Heaton and Johannes Heidecke and Chris Hesse and Alan Hickey and Wade Hickey and Peter Hoeschele and Brandon Houghton and Kenny Hsu and Shengli Hu and Xin Hu and Joost Huizinga and Shantanu Jain and Shawn Jain and Joanne Jang and Angela Jiang and Roger Jiang and Haozhun Jin and Denny Jin and Shino Jomoto and Billie Jonn and Heewoo Jun and Tomer Kaftan and Łukasz Kaiser and Ali Kamali and Ingmar Kanitscheider and Nitish Shirish Keskar and Tabarak Khan and Logan Kilpatrick and Jong Wook Kim and Christina Kim and Yongjik Kim and Jan Hendrik Kirchner and Jamie Kiros and Matt Knight and Daniel Kokotajlo and Łukasz Kondraciuk and Andrew Kondrich and Aris Konstantinidis and Kyle Kosic and Gretchen Krueger and Vishal Kuo and Michael Lampe and Ikai Lan and Teddy Lee and Jan Leike and Jade Leung and Daniel Levy and Chak Ming Li and Rachel Lim and Molly Lin and Stephanie Lin and Mateusz Litwin and Theresa Lopez and Ryan Lowe and Patricia Lue and Anna Makanju and Kim Malfacini and Sam Manning and Todor Markov and Yaniv Markovski and Bianca Martin and Katie Mayer and Andrew Mayne and Bob McGrew and Scott Mayer McKinney and Christine McLeavey and Paul McMillan and Jake McNeil and David Medina and Aalok Mehta and Jacob Menick and Luke Metz and Andrey Mishchenko and Pamela Mishkin and Vinnie Monaco and Evan Morikawa and Daniel Mossing and Tong Mu and Mira Murati and Oleg Murk and David Mély and Ashvin Nair and Reiichiro Nakano and Rajeev Nayak and Arvind Neelakantan and Richard Ngo and Hyeonwoo Noh and Long Ouyang and Cullen O'Keefe and Jakub Pachocki and Alex Paino and Joe Palermo and Ashley Pantuliano and Giambattista Parascandolo and Joel Parish and Emy Parparita and Alex Passos and Mikhail Pavlov and Andrew Peng and Adam Perelman and Filipe de Avila Belbute Peres and Michael Petrov and Henrique Ponde de Oliveira Pinto and Michael and Pokorny and Michelle Pokrass and Vitchyr H. Pong and Tolly Powell and Alethea Power and Boris Power and Elizabeth Proehl and Raul Puri and Alec Radford and Jack Rae and Aditya Ramesh and Cameron Raymond and Francis Real and Kendra Rimbach and Carl Ross and Bob Rotsted and Henri Roussez and Nick Ryder and Mario Saltarelli and Ted Sanders and Shibani Santurkar and Girish Sastry and Heather Schmidt and David Schnurr and John Schulman and Daniel Selsam and Kyla Sheppard and Toki Sherbakov and Jessica Shieh and Sarah Shoker and Pranav Shyam and Szymon Sidor and Eric Sigler and Maddie Simens and Jordan Sitkin and Katarina Slama and Ian Sohl and Benjamin Sokolowsky and Yang Song and Natalie Staudacher and Felipe Petroski Such and Natalie Summers and Ilya Sutskever and Jie Tang and Nikolas Tezak and Madeleine B. Thompson and Phil Tillet and Amin Tootoonchian and Elizabeth Tseng and Preston Tuggle and Nick Turley and Jerry Tworek and Juan Felipe Cerón Uribe and Andrea Vallone and Arun Vijayvergiya and Chelsea Voss and Carroll Wainwright and Justin Jay Wang and Alvin Wang and Ben Wang and Jonathan Ward and Jason Wei and CJ Weinmann and Akila Welihinda and Peter Welinder and Jiayi Weng and Lilian Weng and Matt Wiethoff and Dave Willner and Clemens Winter and Samuel Wolrich and Hannah Wong and Lauren Workman and Sherwin Wu and Jeff Wu and Michael Wu and Kai Xiao and Tao Xu and Sarah Yoo and Kevin Yu and Qiming Yuan and Wojciech Zaremba and Rowan Zellers and Chong Zhang and Marvin Zhang and Shengjia Zhao and Tianhao Zheng and Juntang Zhuang and William Zhuk and Barret Zoph},
      year={2024},
      eprint={2303.08774},
      archivePrefix={arXiv},
      primaryClass={cs.CL},
      url={https://arxiv.org/abs/2303.08774}, 
}

@article{internvl3-2024,
  title={Expanding Performance Boundaries of Open-Source Multimodal Models with Model, Data, and Test-Time Scaling},
  author={Chen, Zhe and Wang, Weiyun and Cao, Yue and Liu, Yangzhou and Gao, Zhangwei and Cui, Erfei and Zhu, Jinguo and Ye, Shenglong and Tian, Hao and Liu, Zhaoyang and others},
  journal={arXiv preprint arXiv:2412.05271},
  year={2024}
}

@article{Oh&Linzen2025,
      title={To Model Human Linguistic Prediction, Make LLMs Les
             Superhuman}, 
      author={Byung-Doh Oh and Tal Linzen},
      year={2025},
      journal={ar{X}iv preprint arXiv:2510.05141},
      eprint={2510.05141},
    archivePrefix = {arXiv},
   primaryClass = {cs.CV},

}

\appendix
\section{Appendix}
Our experiments with open-source LLMs were conducted on 1 NVIDIA Tesla A40 GPU with 48GB RAM. The experiments lasted for about 30 hours.

\end{document}